\documentclass[lettersize,journal]{IEEEtran}
\usepackage{amsmath,amsfonts}
\usepackage{algorithm}
\usepackage{algpseudocode}
\usepackage{array}
\usepackage[caption=false,font=normalsize,labelfont=sf,textfont=sf]{subfig}
\usepackage{textcomp}
\usepackage{stfloats}
\usepackage{url}
\usepackage{verbatim}
\usepackage{graphicx}
\usepackage{cite}
\usepackage{color}
\usepackage{xcolor}
\usepackage{bm}
\usepackage{amsmath}
\usepackage{amssymb}
\usepackage{multirow}
\usepackage{hyperref}
\usepackage{booktabs}
\usepackage{colortbl}
\usepackage{orcidlink} 
\usepackage{bbm}
\definecolor{myblue}{RGB}{0,0,128}
\definecolor{LightRed}{rgb}{1,0.92,0.92}
\definecolor{LightOrange}{rgb}{1,0.95,0.88}
\definecolor{LightYellow}{rgb}{1.0,1.0,0.84}
\definecolor{LightGreen}{rgb}{0.9,1.0,0.88}
\definecolor{LightCyan}{rgb}{0.9,1,1}
\definecolor{LightBlue}{rgb}{0.9,0.94,1}
\definecolor{Gray}{gray}{0.95}
\begin{document}

\title{Adapting Large Multimodal Models to Distribution Shifts: The Role of In-Context Learning}

\author{Guanglin~Zhou\orcidlink{0000-0002-1305-2622}, 
    Zhongyi~Han\orcidlink{0000-0003-2851-193X},
    Shiming~Chen\orcidlink{0000-0001-9633-3392},
    Biwei~Huang,
    Liming~Zhu,
    Salman~Khan\orcidlink{0000-0002-9502-1749},~\IEEEmembership{Senior~Member,~IEEE},
    Xin~Gao\orcidlink{0000-0002-7108-3574},
    Lina~Yao\orcidlink{0000-0002-4149-839X},~\IEEEmembership{Senior~Member,~IEEE}
\IEEEcompsocitemizethanks{
\IEEEcompsocthanksitem Guanglin Zhou is with the University of New South Wales.
Email: jameszhou.ustc@gmail.com
\IEEEcompsocthanksitem Zhongyi Han is with King Abdullah University of Science and Technology. Email: hanzhongyicn@gmail.com
\IEEEcompsocthanksitem Shiming Chen is with Mohamed bin Zayed University of Artificial Intelligence. Email: gchenshiming@gmail.com
\IEEEcompsocthanksitem Biwei Huang is with University of California, San Diego. Email: bih007@ucsd.edu
\IEEEcompsocthanksitem Liming Zhu is with CSIRO's Data61. Email: liming.zhu@data61.csiro.au
\IEEEcompsocthanksitem Salman Khan is with the Mohamed bin Zayed University of Artificial Intelligence and Australian National University. Email: salman.khan@mbzuai.ac.ae
\IEEEcompsocthanksitem Xin Gao is with King Abdullah University of Science and Technology. Email: xin.gao@kaust.edu.sa
\IEEEcompsocthanksitem  Lina Yao is with CSIRO's Data61, the University of New South Wales and Macquarie University. Email: lina.yao@unsw.edu.au
}
}

\markboth{Journal of \LaTeX\ Class Files,~Vol.~14, No.~8, August~2021}%
{Shell \MakeLowercase{\textit{et al.}}: A Sample Article Using IEEEtran.cls for IEEE Journals}


\maketitle

\begin{abstract}
Recent studies indicate that large multimodal models (LMMs) potentially act as general-purpose assistants and are highly robust against different distributions.
Despite this, domain-specific adaptation is still necessary particularly in specialized areas like healthcare.
Due to the impracticality of fine-tuning LMMs given their vast parameter space, this work investigates \textit{in-context learning} (ICL) as an effective alternative for enhancing LMMs' adaptability.
% We find that the success of ICL heavily relies on the choice of demonstration, mirroring challenges seen in large language models but introducing unique complexities for LMMs facing distribution shifts.
Our study proceeds this by evaluating an unsupervised ICL method which selects in-context examples through a nearest example search based on feature similarity. 
We uncover that its effectiveness is limited by the deficiencies of pre-trained vision encoders under distribution shift scenarios, evidenced by their zero-shot capabilities barely outperforming random guesses.
To address these challenges, we propose InvariantSelectPR, a novel method leveraging Class-conditioned Contrastive Invariance (CCI) for more robust demonstration selection.
Specifically, CCI enhances pre-trained vision encoders by improving their discriminative capabilities across different classes and ensuring invariance to domain-specific variations.
This enhancement allows the encoders to effectively identify and retrieve the most informative examples, which are then used to guide LMMs in adapting to new query samples under varying distributions.
Our experiments show that InvariantSelectPR substantially improves the adaptability of LMMs, achieving significant performance gains on benchmark datasets, with a {\color{blue}$34.2\%$}$\uparrow$ accuracy increase in 7-shot on Camelyon17 and {\color{blue}$16.9\%$}$\uparrow$ increase in 7-shot on HAM10000 compared to the baseline zero-shot performance.
Our code will be publicly available at: \url{https://github.com/jameszhou-gl/icl-distribution-shift}.
\end{abstract}


\begin{figure}[tbh]
    \begin{center}
        \includegraphics[width=0.96\linewidth,keepaspectratio]{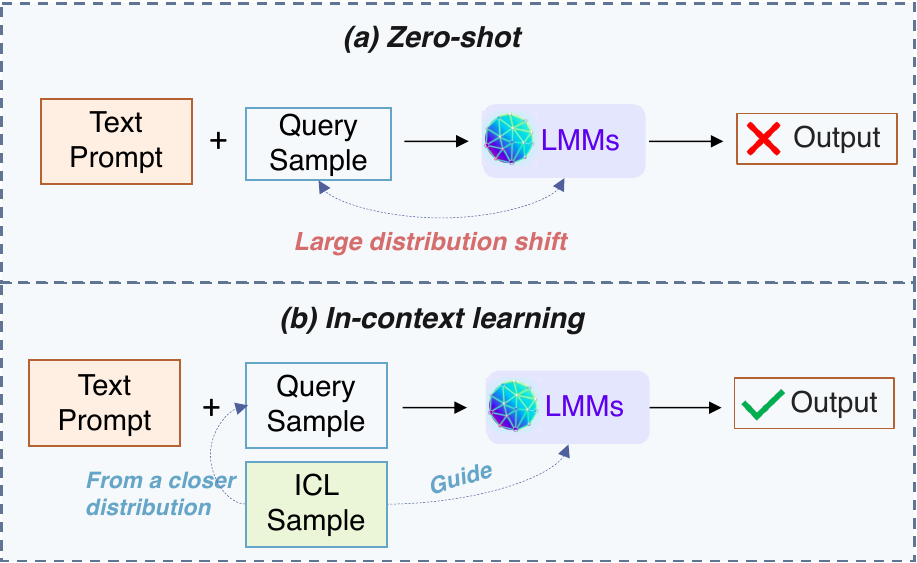}
        \caption{
        Comparative illustration of (a) zero-shot transfer, which relies on LMMs' pre-trained knowledge to respond to queries, potentially leading to a large distribution gap between pre-training data and query samples, 
        and (b) in-context learning (ICL), which introduces an example from a closer distribution with query sample to bridge this gap. 
        This work investigates different retrieval methods for selecting effective ICL examples. 
        }
        \label{546908540649}
    \end{center}
    \vspace{-10pt}
\end{figure}

\section{Introduction}
\IEEEPARstart{M}{achine} learning models are essential in areas such as climate modeling, biomedicine, and autonomous driving, where they need to reliably manage deviations from their training data known as distribution shifts~\cite{park2021reliable,zhou2022domain,zhou2023emerging}. Traditional methods like domain adaptation (DA) and domain generalization (DG) have been somewhat effective but still fall short in addressing these shifts, as confirmed by several empirical studies~\cite{gulrajani2020search,wiles2022a}.
However, the emergence of foundation models, characterized by their extensive and diverse pretraining, offers new possibilities for enhancing adaptability to these challenges~\cite{bommasani2021opportunities,radford2021learning,pmlr-v202-shu23a}. 
Specifically, large multimodal models (LMMs)~\cite{yang2023dawn} such as GPT-4V~\cite{gpt4v}, Claude~\cite{Anthropic2023} and Gemini~\cite{team2023gemini} have shown superior adaptability.
Their zero-shot\footnote{In this study, zero-shot refers to the ability to apply models to new tasks without additional training~\cite{radford2021learning}. It differs from traditional zero-shot learning of generalizing to unseen categories~\cite{chen2022transzero,chen2023evolving}.} capabilities have been found to frequently outperform the performance of traditional fine-tuned models in natural datasets~\cite{han2024how}.

Despite recent advances, domain-specific adaptation remains a significant challenge, especially in healthcare~\cite{han2024how}. 
While LMMs like Google DeepMind's Med-Gemini offer fine-tuned versions for medical tasks~\cite{saab2024capabilities}, their block-box nature and massive parameter sets make traditional fine-tuning impractical for researchers without extensive computational resources. 
This highlights the urgent need for more feasible adaptation techniques. 
\textit{In-context learning} (ICL), which allows models to adapt during inference without parameter adjustments, emerges as a promising alternative~\cite{brown2020language,liu2021makes,min2022rethinking,dong2022survey}. While the effectiveness of ICL is recognized within large language models (LLMs), its application for improving adaptability in LMMs under distribution shifts is less explored.

As depicted in Figure~\ref{546908540649}, we hypothesize that equipping LMMs with context examples that include task-specific information and details about the query sample can substantially enhance their performance.
Our research starts with a thorough evaluation of ICL's capacity to tailor LMMs (\S \ref{718773627961}) to specific domains, particularly healthcare research, where there is potentially large distribution shift through proxy measures (\S \ref{009211896326}).
We discover that the success of ICL heavily depends on the choice of demonstrations.
% , supporting findings from previous LLMs studies~\cite{liu2021makes, min2022rethinking, dong2022survey}.
% Although the significant impact of demonstration selection on ICL performance is not entirely unexpected, the challenge of selecting demonstrations for LMMs under distribution shifts remains unexplored. 
To address this, we re-examine the unsupervised retrieval of in-context examples (\S \ref{939969701961}), TopKNearestPR, traditionally used in LLMs.
This intuitive method uses feature similarity to pinpoint contextually relevant ICL examples~\cite{liu2021makes,zhang2024makes}.

\begin{figure}[htb] % The 'H' specifier forces the figure to be placed exactly here
\centering
\includegraphics[width=0.96\linewidth,keepaspectratio]{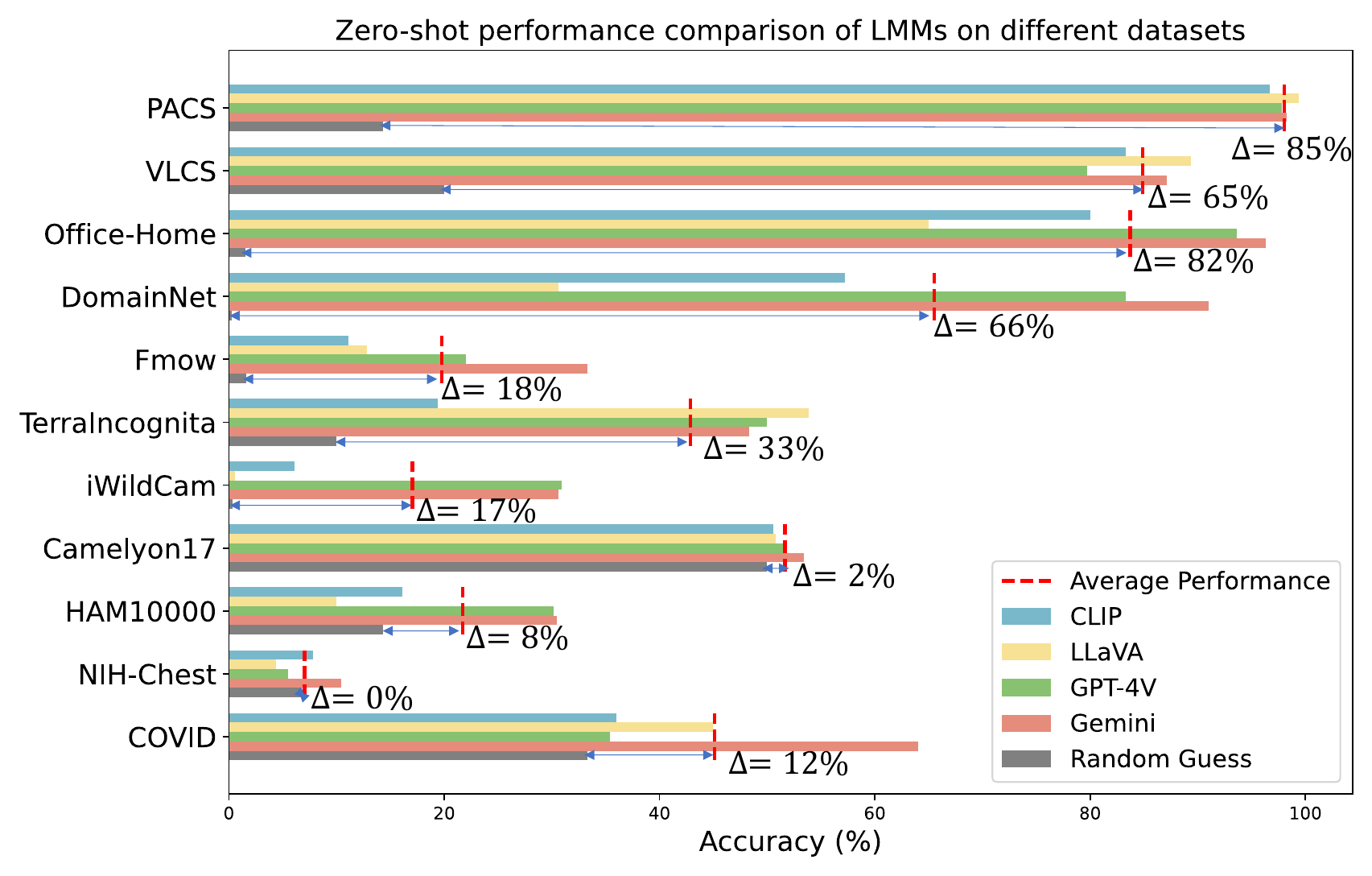}
\caption{A proxy task to evaluate potential distribution shifts in LMMs, illustrating zero-shot performance across various datasets compared to random guessing. 
Red horizontal lines indicate the average performance of LMMs for each dataset, and their minor deviations from random guessing highlight significant shifts, particularly in medical contexts such as Camelyon17, HAM10000, NIH-Chest, and COVID datasets.}
\label{772545618993}
\end{figure}

However, in scenarios of distribution shifts, the TopKNearestPR approach faces considerable challenges when applied to LMMs. 
Notably, using pretrained vision encoders like CLIP-ViT\footnote{\url{https://huggingface.co/openai/clip-vit-base-patch16}}, zero-shot performance often remains at levels comparable to random guessing in specialized domains, as shown in Figure~\ref{772545618993}.
This poor performance reveals a critical limitation in these encoders: they struggle to recognize and adapt to the subtle variations in new distributions, which compromises the reliability of visual feature similarities for selecting effective demonstrations.
To tackle these challenges, we propose InvariantSelectPR, a novel method designed specifically for scenarios involving distribution shifts (\S \ref{180740308016}).
This approach employs Class-conditioned Contrastive Invariance (CCI) to choose demonstrations based on domain-invariant features, which are inherently robust to distributional changes~\cite{zhou2024hcvp}. 
This retriever method is distinctively crafted for distribution shifts, ensuring the resilience of selected in-context examples in varying conditions.
Our empirical results demonstrate that InvariantSelectPR significantly improves the adaptability of LMMs, achieving notable accuracy improvements, \emph{i.e.}, a 34.2\% accuracy improvement in 7-shot on Camelyon17 and a 16.9\% accuracy increase in 7-shot on HAM10000 over the zero-shot baseline.

Our contributions and the key findings are summarized as follows:
(1) To the best of our knowledge, this work takes a first step towards deeply understanding in-context learning as an effective strategy for adapting LMMs to distribution shifts (\S \ref{469902960451}).
(2) We introduce InvariantSelectPR, a novel in-context retrieval framework specifically developed to tackle distribution shifts (\S \ref{996717514819}).
(3) Through extensive experiments on four benchmark datasets (\S \ref{833308195951}), our InvariantSelectPR method shows substantial enhancements over LMMs' zero-shot capabilities.

\section{Related Work}
\subsection{Distribution Shifts} 
The literature on distribution shifts categorizes mitigation approaches into two primary strategies: domain adaptation and domain generalization. Domain adaptation techniques, well-established for scenarios where the target domain is known during training, recalibrate models according to the target data's statistical properties~\cite{ben2006analysis}. These techniques encompass deep transfer learning, which aligns feature distributions between source and target domains \cite{sun2016deep}, unsupervised methods that minimize domain discrepancies \cite{ganin2015unsupervised}.
% and the use of benchmarks such as Office-Home \cite{cite:CVPR2017OfficeHome} and DomainNet \cite{cite:ICCV2019DomainNet}. These benchmarks have propelled advances by introducing more complex and diverse scenarios \cite{han2022towards,han2022learning}. 
In contrast, domain generalization addresses the more daunting challenge of excelling in completely unseen domains. Strategies here include aligning features across multiple source domains \cite{cite:ECCV18CIAN}, separating domain-specific from domain-general features \cite{cite:ICML20CSD,cite:ECCV20DMG}, employing meta-learning for optimization across various domains \cite{cite:AAAI18MLDG,cite:NIPS18MetaReg}, and using data augmentation to mimic domain variability \cite{cite:NIPS2018ADA, cite:CVPR19JiGen}. Recent studies have observed that LMMs have demonstrated exceptional adaptability when dealing with natural datasets but cannot handle the distributions in specialized areas such as healthcare \cite{radford2021learning, han2023well}. This observation motivates this paper's to explore distribution shifts in LMMs through proxy measures and investigate the ICL strategies under distribution shifts.

\subsection{In-Context Learning} 
In-context learning (ICL), particularly defined in GPT-3 \cite{brown2020language}, originated in LLMs for natural language processing (NLP) tasks. ICL is a proven effective paradigm that leverages context augmented with a few examples to enable LLMs to make predictions \cite{dong2022survey, lu2021fantastically, wei2022chain, min2022rethinking, dong2022survey, wei2022emergent, wolf2023fundamental, wies2024learnability, xie2021explanation}. The choice of these in-context examples critically impacts performance, as evidenced by studies demonstrating that selecting nearest neighbors based on sentence encoders can significantly enhance the few-shot capabilities of models like GPT-3 \cite{liu2021makes, wu2024infoprompt, mao2024tuning}. While ICL is established in NLP, it is emerging in visual and multimodal LLMs. The study Flamingo \cite{alayrac2022flamingo} marks the earliest exploration of visual ICL, with subsequent studies validating the importance of example selection in image painting models \cite{bar2022visual, wang2023images, zhang2024makes}, visual understanding~\cite{balazevic2024towards, fang2024explore}, and diffusion models \cite{wang2024context}. Unlike prior work, this paper uniquely focuses on deeply understanding the role of ICL under distribution shifts, taking a first step in this direction.

\begin{figure*}[tb]
\centering
\includegraphics[width=0.96\linewidth,keepaspectratio]{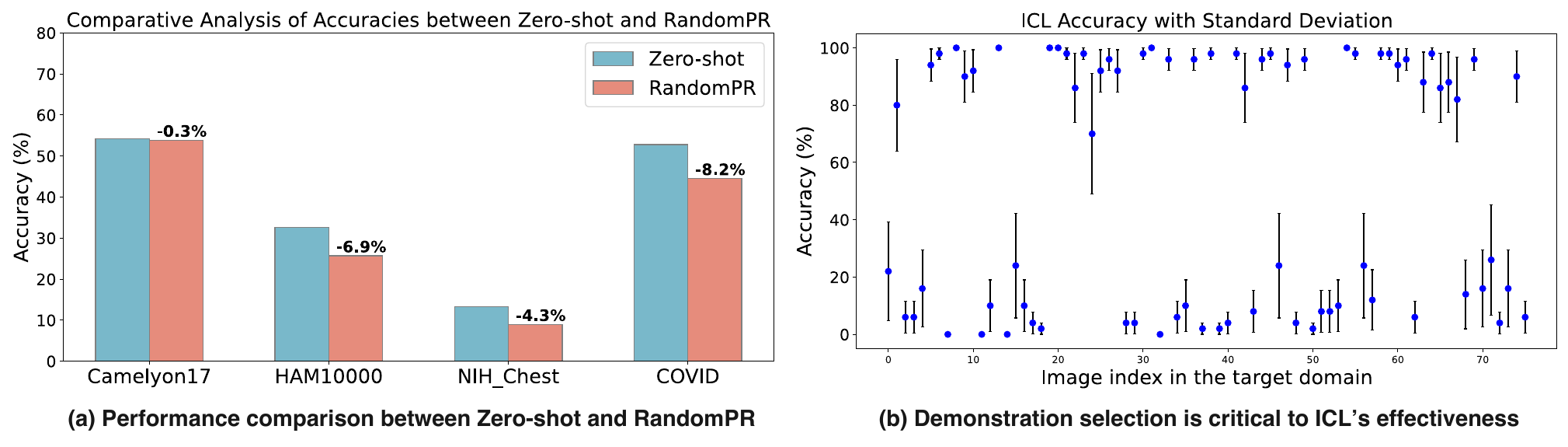}
\caption{ICL Demonstrations under Distribution Shifts: (a) Performance comparison between Zero-shot and RandomPR, illustrating the limitations of random in-context example selection across four datasets, where one-shot RandomPR often underperforms compared to zero-shot. 
(b) Analysis of 77 query samples from the target domain, \texttt{hospital\_3} in Camelyon17, using 50 distinct one-shot examples to examine performance variability. 
Mean values are marked in blue, and variance is represented by black lines, highlighting the significant impact of example selection on model accuracy.
If appropriate in-context samples are chosen, there is a potential for gains up to 40.25\%.}
\label{751719482771}
\end{figure*}

\section{Motivation}
\label{469902960451}

\subsection{Evaluating Distribution Shifts in LMMs Through Proxy Measures}
\label{009211896326}
Distribution shifts traditionally refer to discrepancies between training and test data distributions~\cite{han2022towards,wang2022generalizing}. 
Traditional domain adaptation (DA) and domain generalization (DG) tackle these shifts by fine-tuning models on training data to enhance performance across various yet related test distributions.
Instead, LMMs are engineered to serve as general-purpose assistants~\cite{li2023multimodal}, effectively operating in an efficient way in zero-shot or few-shot modes without the need for parameter adjustments.
Therefore, we conceptualize distribution shifts in LMMs as the gap between the data distributions seen during their pre-training and the data distributions in test scenarios.
This shift is particularly challenging to quantify directly because details of the pre-training data are often not publicly available.

To tackle this, we adopt using a proxy task approach, inspired by similar strategies from related research on data contamination~\cite{bordt2024much,bordt2024elephants}.
This involves comparing the zero-shot performance against a random guessing baseline to infer potential distribution shifts.
Minor deviations from this baseline indicate a potential distribution shift.
When test data features are not well-represented within the LMM's pre-training data, we expect the model to perform only slightly better than random guessing.
Such small performance differences indicate that the model is facing unfamiliar data, highlight a substantial shift in distribution.

To build a suitable benchmark for our study, we initially collected data from eleven public datasets. 
We then applied our proxy method to assess distribution differences, specifically, we evaluated four LMMs across various datasets and measured their zero-shot effectiveness, and calculated these results against the expected performance of random guessing\footnote{We follow the procedures in \url{https://github.com/jameszhou-gl/gpt-4v-distribution-shift}}, as illustrated in Figure~\ref{772545618993}.
Our findings reveal pronounced shifts in medical datasets compared to natural datasets, such as Camelyon17, HAM10000, NIH-Chest, and COVID.
These four datasets serve as our primary benchmarks.
This analysis not only underscores the distribution shifts but also highlights the necessity of adaptive strategies for LMMs, which is the reason that we investigate and design effective ICL methods.

\subsection{ICL Demonstrations under Distribution Shifts}
\label{718773627961}
In this section, we evaluate ICL's capability to enhance LMM adaptability. 
Starting with RandomPR, we randomly select in-context examples from source domain data without relevance to the target task. Our evaluation uses the Gemini model, noted for its zero-shot capabilities, across four medical datasets typically needing domain-specific fine-tuning~\cite{han2024how}. 
We compare one-shot RandomPR with the zero-shot baseline for a preliminary investigation.
According to Figure~\ref{751719482771}(a), while RandomPR presents a slight decrease of 0.3\% on the Camelyon17 dataset, it leads to a substantial performance decline of 4.2\%, 3.7\%, and even 8.2\% on the HAM10000, NIH\_Chest, and COVID datasets respectively.
Despite its conceptual simplicity, our empirical results suggest that random in-context example selection often fails to fulfill the essential requirement for effective model adaptation---providing informative and contextually appropriate demonstrations.   

To unravel the variable efficacy of RandomPR, we conduct an experiment using the Camelyon17 dataset, focusing on 77 query samples from the target domain \texttt{hospital\_3}.
We test the influence of introducing 50 distinct examples from the source domains on the predictions for each query sample.
The results in Figure~\ref{751719482771}(b), display both the mean and standard deviation, indicating significant variability in performance based on the in-context examples used.
Notably, while zero-shot accuracy is 54.55\% ($44/77$), our analysis reveals that up to 73 query samples could be accurately classified with the apt in-context samples, potentially boosting accuracy by 40.25\%. 
Furthermore, the variability observed---such as a mean accuracy of 70\% and a 21\% variance for the 25th query---highlights the varying effects of different ICL examples.
These findings highlight the inconsistencies in RandomPR's performance and underscore the need for advanced methodologies in ICL example selection.

% \vspace{-8pt}
\begin{figure*}[t]
\centering
\includegraphics[width=0.86\linewidth,keepaspectratio]{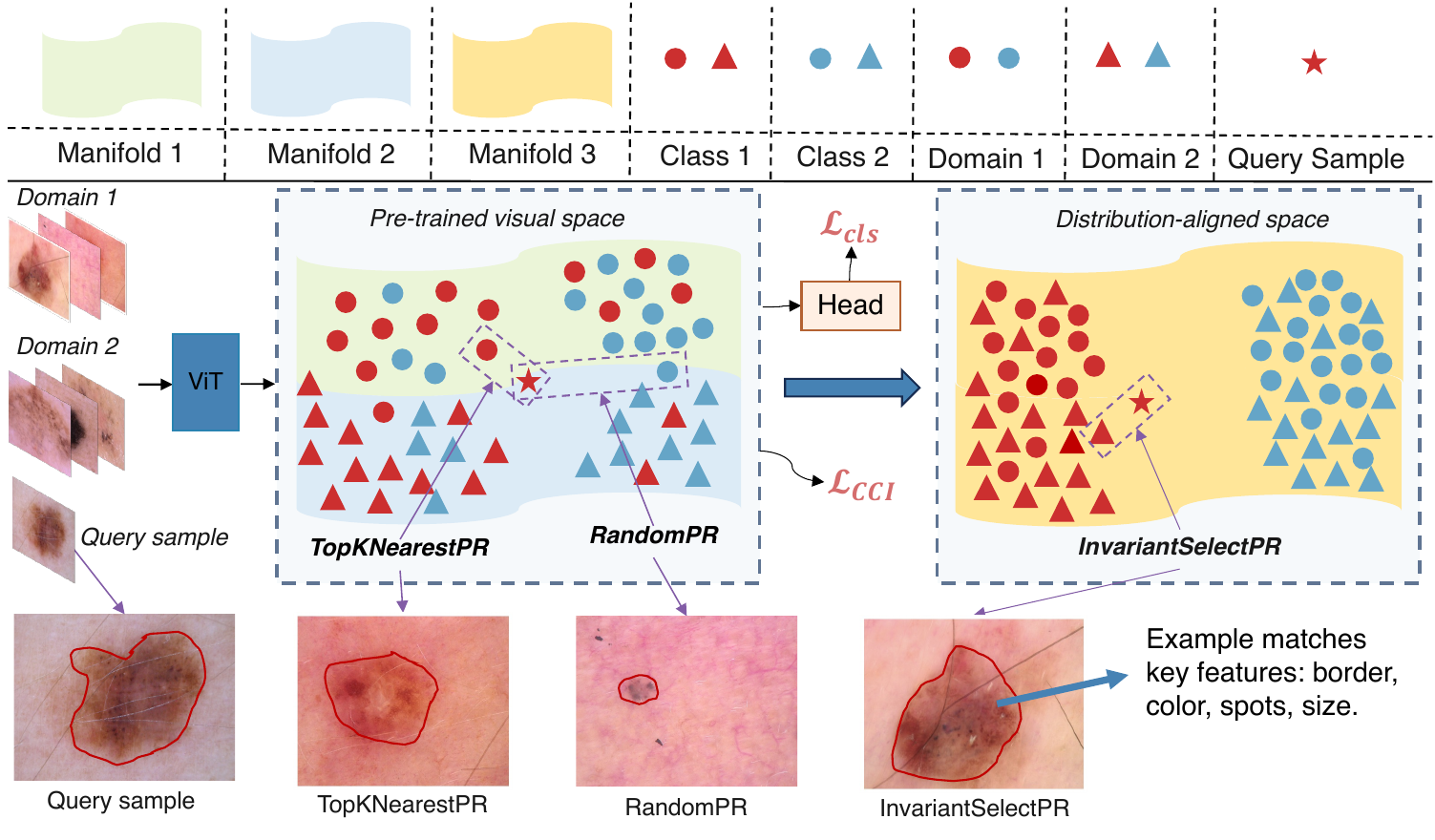}
\caption{Overview of three retrieval methods: RandomPR, TopKNearestPR, and InvariantSelectPR.
RandomPR selects examples without specific criteria, often overlooking informative ones.
TopKNearestPR uses feature similarities for selection, yet struggles with domain-specific tasks where pre-trained encoder features lack sufficient detail. 
In contrast, InvariantSelectPR uses a class-conditioned contrastive invariance (CCI) framework to enhance vision encoders, effectively identifying the most representative samples by focusing on key invariant features.
}
\label{065187225159}
\vspace{-10pt}
\end{figure*}

\section{Methodology}
\label{996717514819}
Upon identifying the limitations of RandomPR, we developed two advanced methods for more effective ICL example selection: TopKNearestPR and InvariantSelectPR, illustrated in Figure~\ref{065187225159}. 
These methods aim to enhance the adaptability of LMMs to distribution shifts through the strategic selection of demonstrative examples.
We detail these selection methods below.

\subsection{TopKNearestPR: Enhancing Context Relevance}
\label{939969701961}
TopKNearestPR adopts an unsupervised strategy to identify in-context examples by measuring the similarity between the feature vectors of a target query image $\bm{x}_q$ and those across $M$ source domains. 
The dataset $\mathcal{S}$ includes domains $\mathcal{S}^i = \{(\bm{x}_{j}^i, \bm{y}_j^i)\}_{j=1}^{n_i}$, where $\bm{x}$ represents feature vectors and $\bm{y}$ is class labels. 
The cosine similarity between the feature vectors from the query image $\bm{x}_q$ and any image $\bm{x}_j^i$ from the dataset, calculated using a pre-trained vision encoder like CLIP-ViT, is given by:
\begin{equation}
    \text{sim}(\bm{x}_q, \bm{x}_j^i) = \frac{\bm{z}(\bm{x}_q) \cdot \bm{z}(\bm{x}_j^i)}{\|\bm{z}(\bm{x}_q)\| \|\bm{z}(\bm{x}_j^i)\|}
\end{equation}
Here, $\bm{z}(\bm{x})$ refers to the feature vector extracted by the encoder. 
The top $K$ images that exhibit the highest similarity to the query are selected using:
\begin{equation}
    \text{top}_K\left(\{\text{sim}(\bm{x}_q, \bm{x}_j^i) : i = 1, \ldots, M; j=1, \ldots, n_i\}\right)
\end{equation}
where $\text{top}_K$ denotes the operation of selecting the indices of the $K$ largest values from the set.
The selected images serve as the in-context examples for the LMMs, aiming to enhance their understanding and performance on analogous tasks without further training.

\subsection{InvariantSelectPR: Tailored for Distribution Shift Adaptation}
\label{180740308016}
TopKNearestPR focuses on relevance by utilizing feature similarities, but its effectiveness can be constrained by the granularity of features from conventional encoders. 
Pretrained vision encoders, such as CLIP-ViT, while robust in general scenarios, often struggle to differentiate effectively in domain-specific tasks.
This limitation manifests as zero-shot performances that are only marginally better than random guesses, leading to the selection of suboptimal in-context examples when relying solely on pre-trained models.
Thus, we propose InvariantSelectPR, a new method designed to enhance robustness across distribution shifts.

\subsubsection{Facilitating Class-conditioned Contrastive Invariance}
InvariantSelectPR is centered around the Class-conditioned Contrastive Invariance (CCI) mechanism, which aims to improve the model's ability to distinguish between classes while maintaining stability across domain-specific variations~\cite{zhou2024hcvp}. 
This is achieved by promoting similarity among instances of the same class from different domains and highlighting differences between classes. 
Using the class token embedding [\text{CLS}], $\bm{x}_N$, from the final vision transformer (ViT) layer, the CCI loss is defined as:

\begin{equation}
\label{656501625535}
\mathcal{L}_{\text{CCI}} = -\mathbb{E}\left[\log \frac{\exp(\bm{z}_N \cdot \bm{z}_{N'} / \tau)}{\sum_{k \neq N}\exp(\bm{z}_N \cdot \bm{z}_{k} / \tau)}\right]
\end{equation}
Here, \(\bm{z}_{N'}\) is a positive sample of \(\bm{z}_N\) from the same class but possibly a different domain, and \(\bm{z}_{k}\) signifies a negative sample of \(\bm{z}_N\) from a different class.
$\tau$ denotes the temperature parameter in contrastive learning~\cite{chen2020simple, zhou2023contrastive}.
This formulation ensures that the learned representations are both discriminative and invariant, crucial for adapting to new distributions.

This approach combines this CCI loss $\mathcal{L}_{\text{CCI}}$ with a classification loss $\mathcal{L}_{\text{cls}}$ to enhance the vision encoder's ability to manage distribution shifts effectively. 
The classification loss uses cross-entropy to align the final class token embedding $\mathbf{x}_N$ with the ground-truth label $\bm{y}$, bolstering the model's discriminative power:
\begin{equation}
\label{656501625537}
\mathcal{L}_{\text{cls}} = -\sum_{i=1}^{C} \bm{y}_i \log(\text{Head}(\bm{x}_N)_i)
\end{equation}
\begin{equation}
\label{656501625536}
\mathcal{L}_{\text{total}} = \mathcal{L}_{\text{cls}} +  \lambda\mathcal{L}_{\text{CCI}}
\end{equation}
where \( C \) represents the total number of classes in the dataset, and $\text{Head}(\cdot)$ is a neural classification head that maps the class token $\bm{x}_N$ to a predicted class probability distribution.
$\lambda$ is a tuning hyper-parameter to control the weight of the CCI loss.

\subsubsection{In-Context Selection Through Enhanced Invariance}
After fine-tuning the vision encoder with the combined losses in Eq. (\ref{656501625536}), we leverage refined features to assess the similarity between the target samples and in-context examples.
By ensuring these similarities reflect both visual resemblance and domain invariance, the k-shot examples with the highest similarity scores are then selected.

\begin{table}[b]
  \caption{Detailed statistics of the datasets used in the experiments.}
  \label{378633656968}
\centering
\resizebox{\columnwidth}{!}{
\begin{tabular}{lccccc}
\toprule
Dataset     & Prediction Task   & \# Domains & \# Classes  & Example Image                                 \\ \midrule
Camelyon17   & Tumor detection               & 5         & 2       &  \includegraphics[height=2cm]{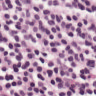} \\ \midrule
HAM10000       & Skin disease classification     & 4         & 7              & \includegraphics[width=2cm]{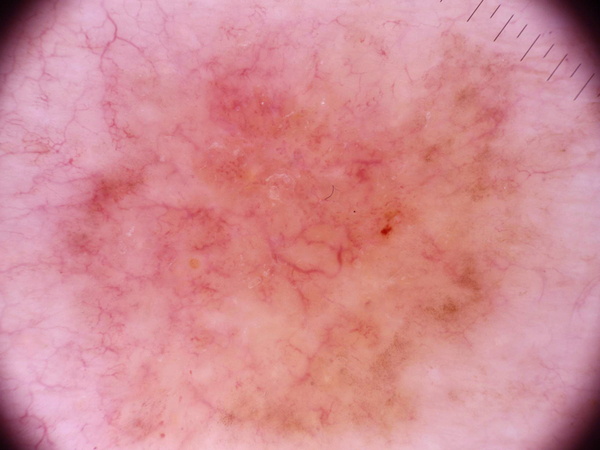} \\ \midrule
NIH\_Chest      & Lung disease diagnosis     & 2         & 15            &  \includegraphics[height=2cm]{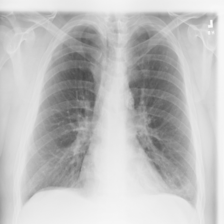}  \\ \midrule
COVID   & Pneumonia type classification  & 2         & 3             &  \includegraphics[height=2cm]{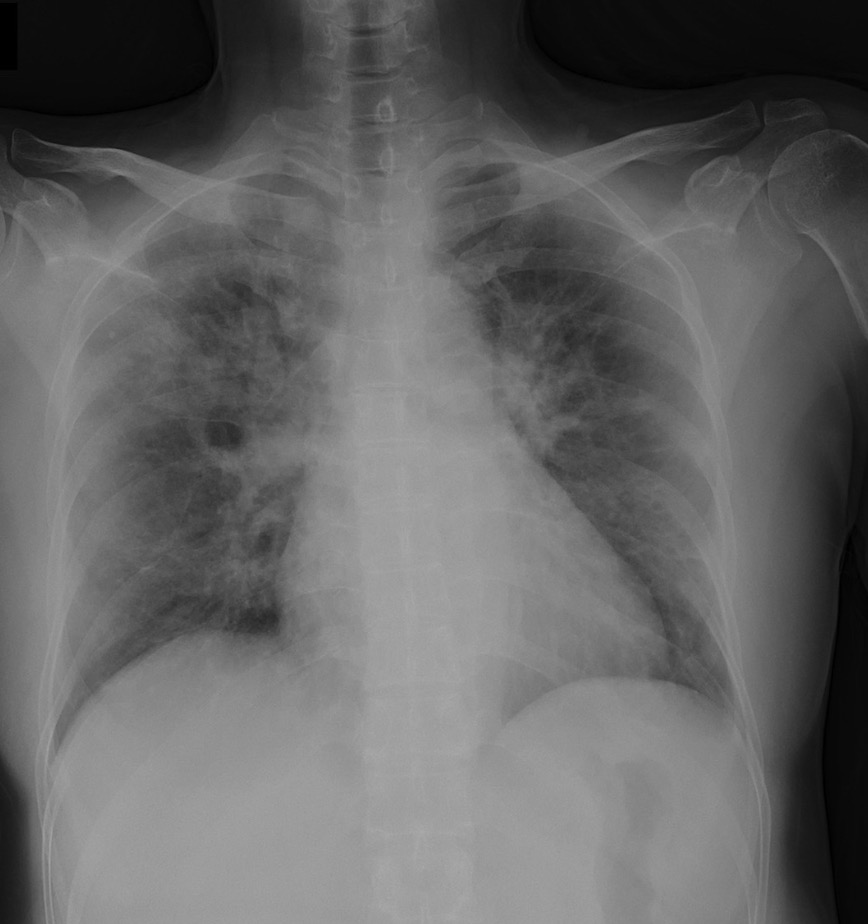}  \\ 
\bottomrule 
\end{tabular}}
\end{table}

\begin{table*}[!t]
\centering
\caption{Performance comparison of three retrieval methods against the zero-shot approach, illustrating accuracy improvements or decreases on Camelyon17 and COVID datasets.
The best and second-best results are marked in \textbf{\color{red}Red} and \textbf{\color{blue}Blue} respectively.
Mean and standard deviation values are calculated over three independent runs.
``$\rightarrow$'' denotes the test scenario.
}
\label{516952833460}
% \vspace{-2mm}
% \newcolumntype{g}{>{\columncolor{Gray}}c}
% \setlength{\tabcolsep}{2.8pt}
% \resizebox{\columnwidth}{!}{
\begin{tabular}{lcccccc|ccc}
\toprule

\multicolumn{1}{l}{\multirow{2}{*}{\textbf{Method}}} & \multicolumn{6}{c|}{\textbf{Camelyon17}} & \multicolumn{3}{c}{\textbf{COVID}} \\
\cmidrule{2-10} & {$\rightarrow$ Hosp0} & $\rightarrow$ Hosp1  & $\rightarrow$ Hosp2 & $\rightarrow$ Hosp3 & $\rightarrow$ Hosp4 & Avg & $\rightarrow$ Sou & $\rightarrow$ Tar & Avg\\
\midrule
Zero-shot &  52.00 &  51.93 &  56.44 &  54.55 &  56.67 &  54.17$\pm$0.5 & 62.19 &  44.19 &  \textbf{\color{blue}52.75$\pm$1.3}\\ \midrule
RandomPR &  50.50 & 53.03 &  54.59 &  53.28 &  58.55 &  53.87$\pm$1.1 &  38.66&  49.86&  44.52$\pm$0.9\\ \midrule
TopKNearestPR &  62.24 &  58.65&  58.62 &  59.65 &  60.15 &  \textbf{\color{blue}59.91$\pm$1.8} &  41.59 &  60.88 &  51.70$\pm$2.7 \\ \midrule
InvariantSelectPR &  60.12 &  63.96 &  62.68 &  63.39 &  64.36 &  \textbf{\color{red}62.77$\pm$1.1} &  39.94 &  67.05 &  \textbf{\color{red}54.15$\pm$1.0} \\ 
\bottomrule
\end{tabular}
\end{table*}

\begin{table*}[!t]
\caption{Performance comparison of three retrieval methods against the zero-shot approach, illustrating accuracy improvements or decreases on HAM10000 and NIH\_Chest datasets.
The best and second-best results are marked in \textbf{\color{red}Red} and \textbf{\color{blue}Blue} respectively.
Mean and standard deviation values are calculated over three independent runs.
``$\rightarrow$'' denotes the test scenario.
}
\label{754290944685}
\centering
% \newcolumntype{g}{>{\columncolor{Gray}}c}
\begin{tabular}{lccccc|ccc}
\toprule
\multicolumn{1}{l}{\multirow{2}{*}{\textbf{Method}}} & \multicolumn{5}{c|}{\textbf{HAM10000}} & \multicolumn{3}{c}{\textbf{NIH\_Chest}} \\
\cmidrule{2-9} & $\rightarrow$RD & $\rightarrow$VMod  & $\rightarrow$VMol & $\rightarrow$VDis & Avg &$\rightarrow$PA&$\rightarrow$AP&Avg\\
\midrule
Zero-shot & 28.54 &  37.39 &  26.42 &  42.22 &  \textbf{\color{blue}32.62$\pm$1.2} &  12.41 &  14.26 &  \textbf{\color{blue}13.31$\pm$0.7} \\ \midrule
RandomPR &  21.48 &  32.81 &  27.88 &  12.64 &  25.71$\pm1.9$ &  7.91 &  10.12 &  8.98$\pm$2.1 \\ \midrule
TopKNearestPR &  23.03 &  33.39 &  31.73 &  28.03 &  28.72$\pm$1.0 &  10.34 &  10.43 &  10.39$\pm$1.0 \\  \midrule
InvariantSelectPR &  38.20 &  49.43 &  38.96 &  25.19 &  \textbf{\color{red}40.91$\pm$1.0} &  13.22 &  13.63 &  \textbf{\color{red}13.42$\pm$0.7} \\ 
\bottomrule
\end{tabular}
\end{table*}

\section{Experiments}
\label{833308195951}
\subsection{Experimental Setup}
\subsubsection{Datasets Overview}
We use four benchmark datasets to explore distribution shifts, particularly emphasizing domain-specific fine-tuning~\cite{han2024how}. 
Camelyon17~\cite{bandi2018detection} features 450,000 patches from breast cancer images across five hospitals. 
HAM10000~\cite{tschandl2018ham10000} offers dermatoscopic images critical for skin cancer detection. 
The NIH\_Chest dataset~\cite{wang2017chestx} includes over 112,000 X-ray images annotated for thoracic diseases. 
The COVID dataset~\cite{han2021semi} provides diverse pneumonia detection data, including COVID-19 cases, from various hospitals. 
We analyze a practical subset, \texttt{random\_1}, with 450 samples\footnote{Available at \url{https://github.com/jameszhou-gl/gpt-4v-distribution-shift}}.
Figure~\ref{378633656968} presents detailed statistics of the datasets.

\subsubsection{Implementation Details} 
\begin{figure}[t]
\centering
\includegraphics[width=0.5\textwidth]{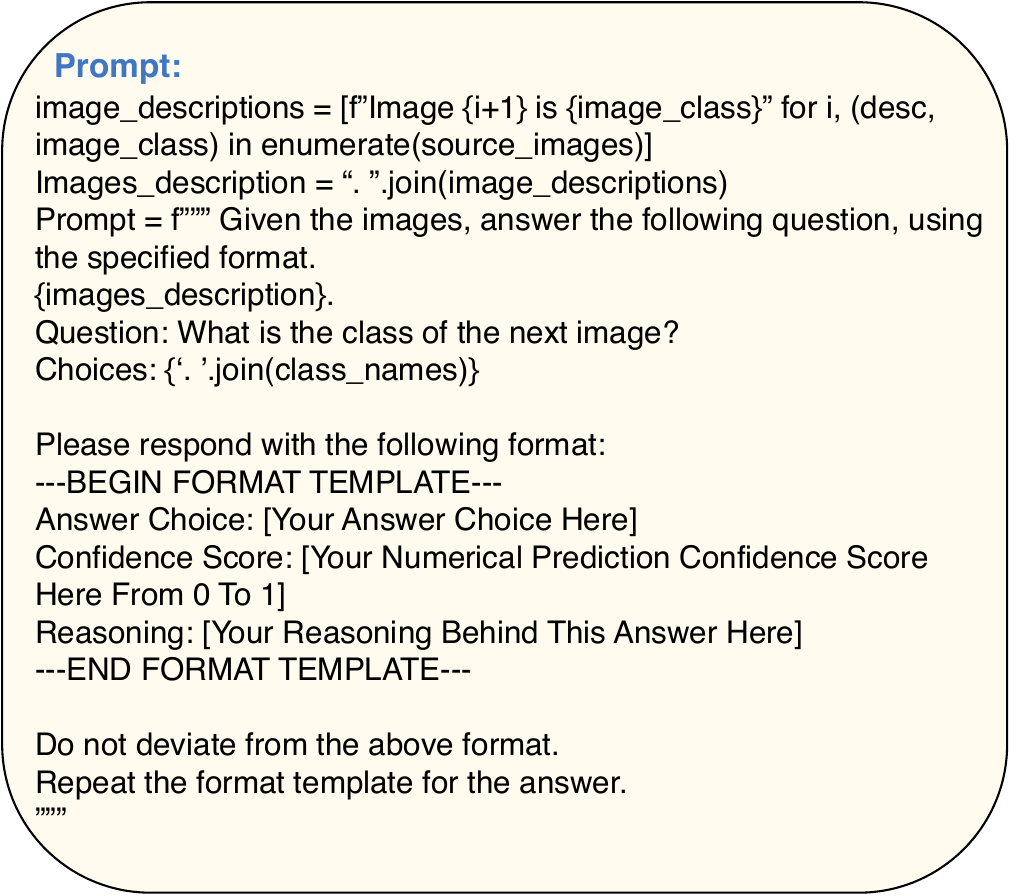}
\caption{Basic prompt template in all ICL experiments.}
\label{538081895648}
\end{figure}

We compare three retrieval methods---RandomPR, TopKNearestPR and InvariantSelectPR---against the baseline zero-shot capability.
We employ \texttt{vit\_large\_patch14\_224\_clip\_laion2b} configuration from the timm library, exploring variations in backbone configurations further in \S \ref{538943692504}.
The Gemini model is employed as the primary LMM due to its superior zero-shot performance across varied datasets~\cite{han2024how} and its stable log-linear improvement in performance with an increasing number of ICL examples, as observed in a concurrent study~\cite{jiang2024manyshot}.
Our main results (\S \ref{294114396348}) focus on one-shot performance, with additional insights on the impact of different numbers of shots in \S \ref{023626146559}.
We also include other leading LMMs, such as GPT-4V and Claude~\cite{Anthropic2023}, in our extended analysis in \S \ref{132518591439}.
% We employ the \texttt{vit\_large\_patch14\_224\_clip\_laion2b} configuration from the timm library\footnote{\url{https://github.com/huggingface/pytorch-image-models/blob/main/timm/models/vision_transformer.py}}.
We utilize the following basic prompt template in all ICL experiments as shown in Figure~\ref{538081895648}.

\subsubsection{Training Protocol} We train InvariantSelectPR for 100 epoches using the AdamW optimizer~\cite{Loshchilov2017DecoupledWD}, with a learning rate of 1$e$-5 and a weight decay of 0.01.
For clarity, both the temperature parameter $\tau$ and the loss weight $\lambda$ are fixed at 1.0.
Each dataset is specifically fine-tuned to optimize the vision encoder for its respective domains.
Experiments are conducted on a Linux server equipped with an Intel Xeon CPU, NVIDIA A5000 and V100 GPUs.

\subsection{Main Results}
\label{294114396348}
In Tables \ref{516952833460} and \ref{754290944685}, our analysis of four benchmark datasets provides a detailed examination of how different ICL methods perform under distribution shifts. 
The zero-shot approach highlights the inherent ability of LMMs to adapt to new domains without retraining. 
However, the effectiveness of this adaptability varies significantly with different ICL strategies.
The RandomPR strategy, which employs a stochastic method for selecting in-context examples, yields inconsistent results. 
For instance, on the Camelyon17 dataset, it leads to a slight decrease of 0.3\% in accuracy, but it largely underperforms on the COVID, HAM10000, and NIH\_Chest datasets, with accuracy decreases of 8.2\%, 4.2\%, and 3.7\%, respectively.
This highlights the unpredictable performance of RandomPR across different conditions.
Conversely, TopKNearestPR uses a pre-trained vision encoder to identify feature similarities for example selection, leading to a 5.74\% improvement on the Camelyon17, which demonstrates the benefits of a more targeted approach in example selection. 
Despite this success, the method sees declines of 3.9\% and 2.9\% on the HAM10000 and NIH\_Chest datasets, respectively, indicating a lack of consistent performance across all test scenarios.
The most effective strategy, InvariantSelectPR, consistently outperforms other methods, significantly exceeding the zero-shot baseline across all datasets, especially achieving remarkable gains of 8.3\% on HAM10000 and 8.6\% on Camelyon17.
These results underscore the importance of advanced in-context example selection techniques in adapting LMMs to distribution shifts.
Despite notable gains, the improvements with InvariantSelectPR on NIH\_Chest and COVID are modest.
In Figure~\ref{705379351913}, the incremental improvements by InvariantSelectPR align with those from fine-tuned encoders, which generally surpass the fine-tuning approach by 1\% to 6\%.
This suggests that when fine-tuning itself is minimally effective, ICL strategies yield limited enhancements.
Future research thus focuses on more sophisticated methods to enhance invariance, beyond fundamental domain-invariance in this work.

\subsection{Ablation Study}
\label{941426154964}

To assess the impact of enhanced invariance on model adaptability, we conduct an ablation study focusing on the Gemini model's one-shot performance. 
This study compares three configurations: a baseline using TopKNearestPR, the baseline only with $\mathcal{L}_{\text{cls}}$, and the full InvariantSelectPR that incorporates both $\mathcal{L}_{\text{cls}}$ and $\mathcal{L}_{\text{CCI}}$.
Table~\ref{128509768508} displays incremental performance gains across datasets with the successive additions of $\mathcal{L}_{\text{cls}}$ and $\mathcal{L}_{\text{CCI}}$. 
The addition of $\mathcal{L}_{\text{cls}}$ alone leads to a modest increase in performance by $1.77\%$.
However, when incorporated with $\mathcal{L}_{\text{CCI}}$, there is a more substantial performance boost of $4.01\%$, confirming the effectiveness of CCI loss in improving the models' adaptability. 
\begin{table}[!h]
\caption{Ablation study on loss terms. The baseline is TopKNearestPR.}
\label{128509768508}
\centering
% \vspace{-2mm}
% \newcolumntype{g}{>{\columncolor{Gray}}c}
%  \renewcommand{\arraystretch}{0.9}
%  \setlength{\tabcolsep}{6.1pt}
  \resizebox{\columnwidth}{!}{
\begin{tabular}{lccccc}
\toprule
Configurations& CA & CO & HA & NI & Avg \\ \midrule
baseline &  61.96 & 54.22 & 29.84 & 10.49 & 39.13  \\ \midrule
baseline+$\mathcal{L}_{cls}$ (w/o CCI) &  61.59 & 52.00 & 38.93 & 11.11 & 40.90  \\ \midrule
baseline+$\mathcal{L}_{cls}$+$\mathcal{L}_{CCI}$ (full)  & \textbf{63.90} & \textbf{54.44} & \textbf{41.56} & \textbf{12.67} & \textbf{43.14} \\
\bottomrule
\end{tabular}}
\end{table}

\subsection{In-depth Analysis}
\subsubsection{ICL vs. Traditional Supervised Finetuning (SFT)}
\label{689906963865}
\begin{figure}[htb] % The 'H' specifier forces the figure to be placed exactly here
\centering
\includegraphics[width=0.5\textwidth]{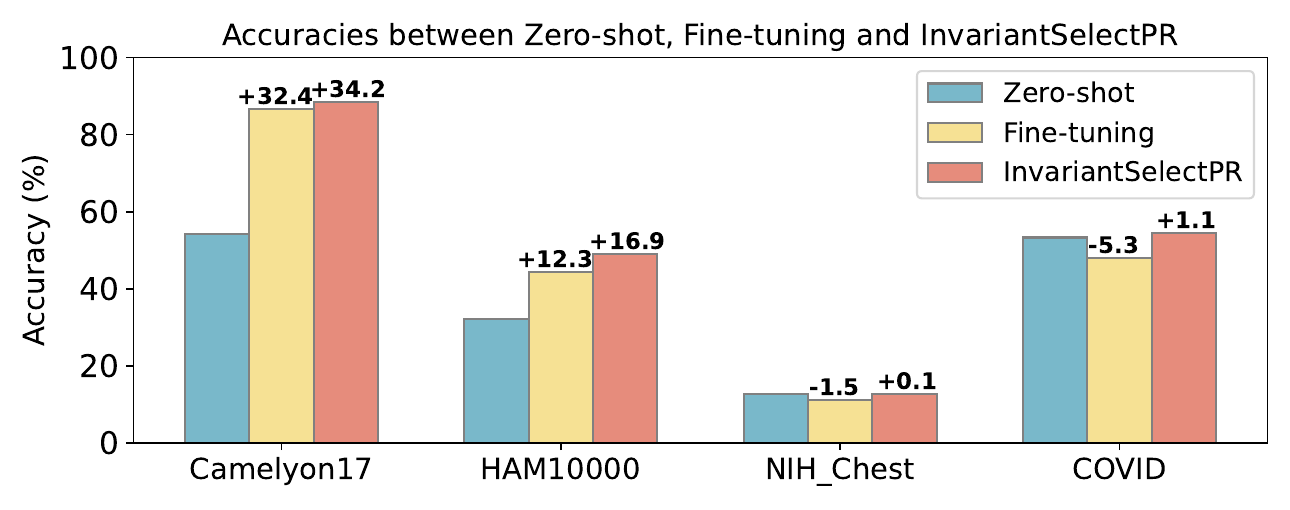}
\vspace{-3mm}
\caption{Comparative accuracies between Zero-shot, Supervised Finetuning, and InvariantSelectPR methods across various datasets, illustrating the superior performance of InvariantSelectPR over both zero-shot and supervised fine-tuning.}
\label{705379351913}
% \vspace{-10pt}
\end{figure}

Recent advances in LMMs demonstrate their impressive zero-shot generalization, often outperforming fine-tuned models in natural distribution shifts~\cite{han2024how}. 
This raises questions about whether LMMs with in-context learning, can exceed fine-tuned model performance in scientific datasets, traditionally reliant on domain-specific fine-tuning.
We assess our ICL strategy against traditional SFT to explore this.
For SFT, we focus on maintaining domain invariance and targeting the class prediction objective, similar to Eq. (\ref{656501625536}). 
We fine-tune a CLIP-ViT on source domain data and then apply it to predict outcomes on target examples.
This comparison directly measures the effectiveness of ICL versus conventional SFT.
For InvariantSelectPR, we choose ICL examples ranging from one to seven and report the best accuracy.
Figure~\ref{705379351913} displays the comparative performance across four datasets. 
SFT demonstrates a substantial improvement over zero-shot capabilities with accuracy improvements of 32.4\%, and 12.3\% on Camelyon17 and HAM10000, but underperforms 1.5\% and 5.3\% on NIH\_Chest and COVID.
In contrast, our proposed InvariantSelectPR with few-shot examples, consistently exceeds SFT, with gains of 1.4\%, 4.4\%, 1.6\%, and 6.4\% in the same datasets.

\subsubsection{Backbone Evaluation}
\label{538943692504}
We evaluate the impact of different vision encoder backbones on the effectiveness of our InvariantSelectPR method compared to TopKNearestPR.
This includes ViT models pretrained on ImageNet-21K
and trained with the self-supervised DINO method on ImageNet-1K.
Table~\ref{526674382045} shows that InvariantSelectPR consistently outperforms TopKNearestPR across datasets, with an average accuracy of 42.9\% versus 39.7\%.
This underscores the limitations of relying solely on pretrained visual similarity for selecting meaningful in-context examples. 
InvariantSelectPR also demonstrates more consistent performance, with less deviation from mean accuracy (under 1\%) compared to TopKNearestPR (nearly 2\%).
An important observation is the enhanced performance of backbones utilizing self-supervised or contrastive learning methods, supporting the effectiveness of self-supervised learning in capturing generalizable features that contribute to more robust ICL performance, as suggested in studies~\cite{chen2020simple,radford2021learning,wang2023scientific}.

\begin{table}[!ht]
\caption{Performance comparison of ICL methods with different vision encoder backbones.}
\label{526674382045}
\centering
% \vspace{-2mm}
% \newcolumntype{g}{>{\columncolor{Gray}}c}
%  \renewcommand{\arraystretch}{0.86}
%  \setlength{\tabcolsep}{4.0pt}
  \resizebox{\columnwidth}{!}{
\begin{tabular}{ccccccc}
\toprule
Methods&Backbones&CA & CO & HA & NI & Avg \\ \midrule
\multirow{3}{*}{TopKNearestPR} & vit-l/14-clip & 61.96 & 54.22 & 29.84 & 10.49 & 39.13  \\
& vit-l/16-in21k & 60.45 & 49.78 & 30.80 & 10.24 & 37.82\\
& vit-b/16-dino & 63.72 & 59.68 & 32.74 & 10.89 & 41.76\\ \midrule
\multirow{3}{*}{InvariantSelectPR} & vit-l/14-clip & 63.90 & 54.44 & 41.56 & 12.67 & 43.14 \\
& vit-l/16-in21k & 61.76 & 52.78 & 38.15 & 12.89 & 41.40\\
& vit-b/16-dino & 64.48 & 53.72 & 44.55 & 11.36 & 43.53\\ 
\bottomrule
\end{tabular}}
% \vspace{-10pt}
\end{table}

\subsubsection{ICL Examples with Various Shots}
\label{023626146559}
\begin{figure}[ht] % The 'H' specifier forces the figure to be placed exactly here
\centering
\includegraphics[width=0.99\linewidth,keepaspectratio]{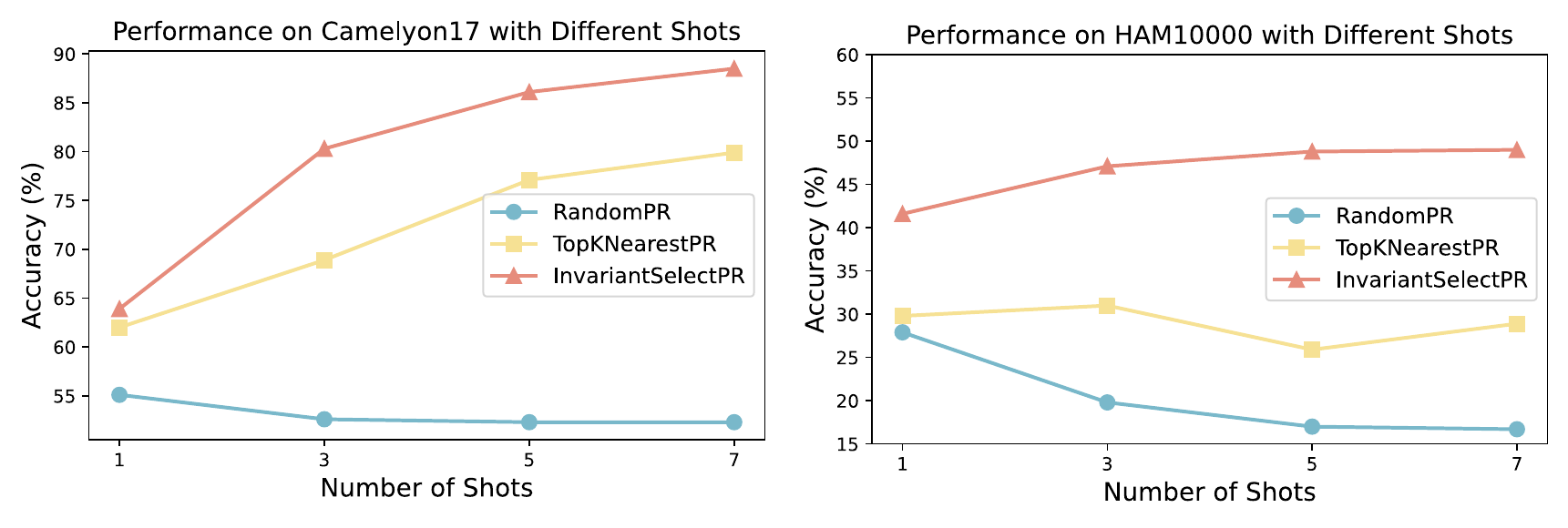}
\caption{Performance comparison with varying numbers of ICL examples (shots) on Camelyon17 and HAM10000 datasets.}
\label{409975459943}
% \vspace{-10pt}
\end{figure}

To assess the impact of the number of ICL examples, we perform an empirical study using the Camelyon17 and HAM10000 datasets, varying the number of shots from 1 to 7 for each dataset in Figure~\ref{409975459943}. 
This analysis reveals that increasing the number of shots leads to a decrease in the performance of the RandomPR method, implying that additional examples might introduce unhelpful information.
In contrast, the TopKNearestPR method typically improves with more shots but shows a decline in performance when moving from 3-shot to 5-shot on the HAM10000 dataset, suggesting potential issues with example selection or redundancy. 
On the other hand, our InvariantSelectPR method consistently improves performance as the number of shots increases, demonstrating its effectiveness in utilizing information from source domains. 
Notably, this method achieves a performance boost of approximately 24.6\% when the shot count increases from 1 to 7 on Camelyon17.

\subsubsection{Evaluation Across Different LMMs}
\label{132518591439}
\begin{figure}[h!] % The 'H' specifier forces the figure to be placed exactly here
\centering
\includegraphics[width=0.5\textwidth]{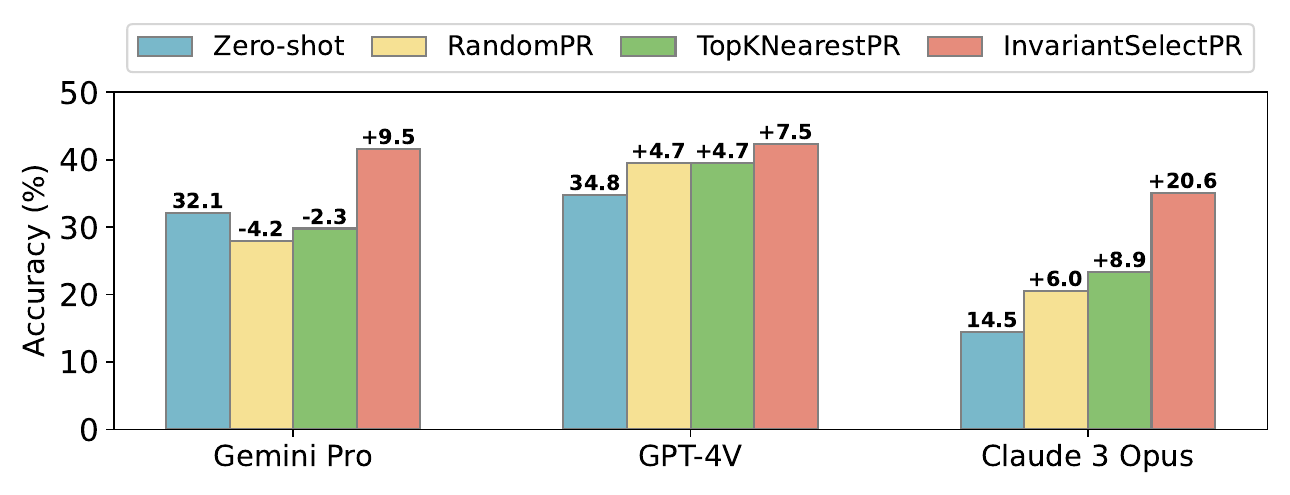}
% \vspace{-4mm}
\caption{Performance comparison of Zero-shot, RandomPR, TopKNearestPR, and InvariantSelectPR on the HAM10000 dataset across three LMMs, demonstrating the impact of one-shot demonstrations.}
\label{080337204301}
\end{figure}

The open-source LMMs like IDEFICS~\cite{laurenccon2024obelics} and OpenFlamingo~\cite{anas_awadalla_2023_7733589} primarily focus on text and ignore the input signal of images~\cite{bertini2024makes}.
Furthermore, these LMMs lack instruction-following ability to choose the response from the answer list.
Thus, we use three proprietary LMMs in this comparative analysis:
Gemini Pro, GPT-4V, and Claude 3 Opus~\cite{Claude_3}. 
Due to the high computational demands and associated costs of GPT-4V and Claude 3 Opus, we limit our testing to a single dataset, HAM10000, and perform a one-shot evaluation.
Figure~\ref{080337204301} demonstrates that InvariantSelectPR consistently outperforms other methods across all three LMMs.
This method not only exceeds baseline zero-shot performance but also significantly enhances adaptability.
Both GPT-4V and Claude 3 Opus exhibit substantial improvements using all ICL methods over their zero-shot capabilities, suggesting that ICL can effectively boost the adaptability of LMMs.
This analysis highlights the capacity of InvariantSelectPR to leverage domain-invariant features to enhance LMMs performance under variable conditions.

\subsubsection{Distance Metric Evaluation}
\label{596370695944}
We examine the effects of employing various distance metrics, including Cosine, Euclidean, and Manhattan, within TopKNearestPR and InvariantSelectPR methods, as illustrated in Table~\ref{060509061776}.
Our findings indicate that the InvariantSelectPR method consistently achieves higher performance than TopKNearestPR across all metrics tested on both the Camelyon17 and HAM10000 datasets. 

\begin{table}[h]
\caption{Performance comparison using different distance metrics.}
\label{060509061776}
\centering
% \newcolumntype{g}{>{\columncolor{Gray}}c}
% \setlength{\tabcolsep}{4.9pt}
 \resizebox{\columnwidth}{!}{
\begin{tabular}{ccccc}
\toprule
{Method} 
& {Cosine} & {Euclidean} & {Manhattan} & Avg \\ \midrule
& \multicolumn{4}{>{\columncolor{LightBlue}}c}{\textit{Camelyon17}} \\ 
TopKNearestPR & 61.96 & 60.77 & 59.18 & 60.64 \\
InvariantSelectPR & 63.90 & 61.09 & 62.70 & 62.56  \\ \midrule
& \multicolumn{4}{>{\columncolor{LightBlue}}c}{\textit{HAM10000}} \\ 
TopKNearestPR & 29.84 & 29.46 & 30.80 & 30.03 \\
InvariantSelectPR & 41.56 & 37.22 & 38.93 & 39.24  \\ 
\bottomrule
\end{tabular}}
\end{table}

% \begin{table*}[h]
% \vspace{-2mm}
% \caption{Performance comparison using different distance metrics.}
% \label{060509061776}
% \centering
% \newcolumntype{g}{>{\columncolor{Gray}}c}
% \begin{tabular}{c|cccg|cccg}
% \toprule
% \multirow{2}{*}{Method} & \multicolumn{4}{c|}{\textbf{Camelyon17}} & \multicolumn{4}{c}{\textbf{HAM10000}} \\
% & {Cosine} & {Euclidean} & {Manhattan} & \textbf{Avg} & {Cosine} & {Euclidean} & {Manhattan} & \textbf{Avg} \\ \midrule
% TopKNearestPR & 61.96 & 60.77 & 59.18 & 60.64 & 29.84 & 29.46 & 30.80 & 30.03\\
% InvariantSelectPR & 63.90 & 61.09 & 62.70 & 62.56 & 41.56 & 37.22 & 38.93 & 39.24 \\ \bottomrule
% \end{tabular}
% \end{table*}

\subsubsection{Computational Efficiency}
Table~\ref{861612844362} illustrates the trade-off between computational cost and accuracy improvement. 
While InvariantSelectPR incurs a slightly higher inference time and GPU usage than the zero-shot baseline but offers an 8.60\% accuracy improvement. 
The increased cost is due to the model loading and similarity calculation. 
InvariantSelectPR' lower inference time compared to TopKNearestPR is because it loads the vision encoder once per environment instead of for each target sample.
Future work will focus on optimizing these steps to reduce inference time while maintaining accuracy gains.

\begin{table}[!ht]
\caption{Performance comparison of different one-shot ICL methods on the Camelyon17 dataset, in terms of inference time, GPU usage, and accuracy improvement over the zero-shot baseline.}
\label{861612844362}
\centering
% \newcolumntype{g}{>{\columncolor{Gray}}c}
%  \renewcommand{\arraystretch}{0.9}
%  \setlength{\tabcolsep}{4.9pt}
 \resizebox{\columnwidth}{!}{
\begin{tabular}{lccc}
\toprule
Method & Time (s/query) & GPU (GB) & Acc Gains \\ \midrule
Zero-shot & 5.23  & - & -  \\ 
RandomPR &  5.35  & - & -0.30\% \\ 
TopKNearestPR  & 15.52 & 2.41 & +5.74\%  \\ 
InvariantSelectPR & 11.79 & 3.55 & +8.60\% \\
\bottomrule
\end{tabular}}
\end{table}

\subsubsection{t-SNE Visualizations of Visual Features}
\begin{figure*}[ht] % The 'H' specifier forces the figure to be placed exactly here
\centering
\includegraphics[width=0.8\linewidth,keepaspectratio]{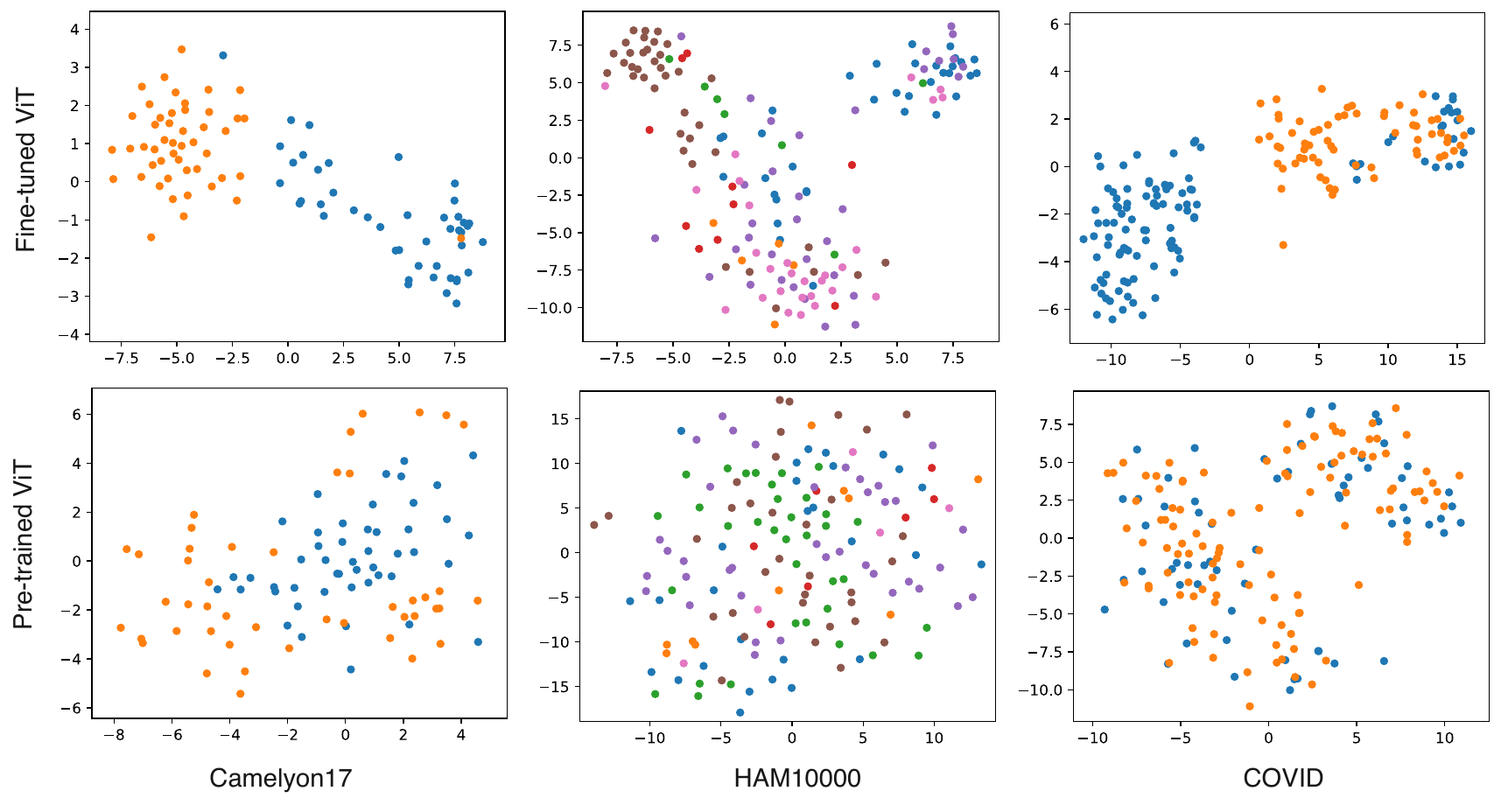}
\caption{t-SNE visualizations of visual features from the target domain for three datasets. The lower row shows features extracted using the pretrained ViT model, while the upper row shows features extracted using the our fine-tuned ViT model.}
\label{044348484546}
\end{figure*}

In this section, we presnet t-SNE~\cite{van2008visualizing} visualizations of the visual features to illustrate how class-conditioned contrastive invariance (CCI) contributes to domain invariance and discriminative capabilities.
The visualizations are based on the original vision encoder and our fine-tuned vision encoder on three datasets, as shown in Figure~\ref{044348484546}.
The visual features are extracted using both the pretrained and fine-tuned ViT models.
The t-SNE plots are created to highlight the clustering behavior of the features from the target domain.
By comparing the plots, we can visually assess the impact of the fine-tuning with CCI on the separation of different classes and the compactness of feature clusters.

These t-SNE visualizations, particularly for the Camelyon17 dataset, clearly demonstrate that fine-tuning with class-conditioned contrastive invariance (CCI) significantly enhances the model’s ability to generalize across unseen domains. The fine-tuning process improves discriminative power by better aligning feature representations with class labels. 
This visualization underscores the critical importance of incorporating CCI to refine the vision encoder. 
By enhancing the alignment of feature representations with their corresponding classes, CCI contributes to a more robust and domain-invariant model. 
Furthermore, this refined vision encoder facilitates the selection of in-context learning (ICL) examples, enabling large multimodal models to adapt more effectively.

% \vspace{-8pt}
\section{Conclusion}
We investigated the efficacy of in-context learning (ICL) to improve the adaptability of LMMs to distribution shifts through our novel ICL approach, InvariantSelectPR. 
This method not only outperforms standard zero-shot capabilities but also exceeds other methods like RandomPR and TopKNearestPR in handling domain-specific shifts. 
Evaluations across four datasets confirmed that InvariantSelectPR enhances LMM adaptability by optimally selecting demonstrative examples. 
Our study offers insights for future work on distribution shifts in foundation models.

Our study has several limitations for further improvements. 
Firstly, we confined our analysis to a small selection of benchmark datasets and relied exclusively on commercial and proprietary models, such as Gemini Pro, GPT-4V, and Claude 3 Opus. 
The limited availability of comprehensive documentation for these models constrains our understanding of their pre-training data, architecture, and inherent biases. 
This is critical as some broadly used open-source LMMs can not effectively understand multiple images like Flamingo~\cite{alayrac2022flamingo} and simultaneously follow instructions like LLaVA~\cite{liu2023visual}, necessitating the use of commercial models. 
Additionally, the substantial financial and computational resources required to access these proprietary models may restrict further validation and analysis.
Secondly, our empirical tests involved just 450 samples, which, despite prior research suggesting stability ranging from 180 to 1800 cases~\cite{han2024how}, might not reveal scalability issues or subtle biases in larger datasets.
Thirdly, the prevalence of numerous domains in healthcare~\cite{yang2023manydg} and scientific research~\cite{ji2022drugood} presents potential challenges in scaling our method.

% \textbf{Boarder Impacts.} \quad This work introduces a novel ICL method to enhance the adaptability of LMMs to distribution shifts, particularly in scientific and healthcare domains. 
% While this method promises improved model accuracy and reliability, its misuse could amplify biases or yield unreliable outputs from LMMs, leading to negative consequences in these critical fields.

\bibliographystyle{IEEEtran}
\bibliography{tmm}

 

% \end{IEEEbiography}



\vfill

\end{document}